\documentclass[11pt]{article}

\PassOptionsToPackage{x11names,table}{xcolor}
\usepackage[final]{acl}

\usepackage{times}
\usepackage{latexsym}

\usepackage{minitoc}
\usepackage{titletoc}
\usepackage{xurl}
\usepackage{hyperref}
\usepackage[T1]{fontenc}

\usepackage[utf8]{inputenc}

\usepackage{microtype}

\usepackage{inconsolata}

\usepackage{graphicx}
\usepackage{amsmath,amssymb,amsfonts}
\usepackage{xcolor}

\usepackage{enumitem}
\usepackage{booktabs}
\usepackage{multirow}
\usepackage{xcolor}

\usepackage{subcaption}
\definecolor{EyeProtectGreen}{HTML}{00693e}

\hypersetup{
    colorlinks=true,
    linkcolor=EyeProtectGreen,
    citecolor=EyeProtectGreen,
    filecolor=EyeProtectGreen,      
    urlcolor=EyeProtectGreen,
    }
\title{Which Tokens Should SFT Actually Learn? A Token-Trimming Perspective on Mathematical Reasoning}

\author{
  \textbf{Yaning Jia\textsuperscript{1}},
  \textbf{Chunhui Zhang\textsuperscript{1}},
  \textbf{Wenxuan Xu\textsuperscript{1}},
  \textbf{Xingjian Diao\textsuperscript{1}}
  \\
  \textbf{Xiaoyuan Wang\textsuperscript{2}},
  \textbf{Soroush Vosoughi\textsuperscript{1}}\thanks{Correspondence to \textit{soroush.vosoughi@dartmouth.edu}} \\
  \textsuperscript{1}Dartmouth College
  \qquad
  \textsuperscript{2}Carnegie Mellon University
  \\
  \small{
    \texttt{yaning.jia.gr@dartmouth.edu}
    \qquad
    \texttt{soroush.vosoughi@dartmouth.edu}
  }
}

\begin{document}
\maketitle
\begin{abstract}
Supervised fine-tuning (SFT) applies a uniform cross-entropy loss to 
all target tokens, even though different tokens provide unequal 
learning signals for mathematical reasoning. This uniform treatment 
can over-sharpen already mastered tokens while amplifying learning 
pressure on uncertain, low-confidence tokens, leading to suboptimal 
training dynamics. We propose \textbf{Trimmed Logit-Gap SFT 
(TrimSFT)}, a simple token-level reweighting method that scales the SFT 
loss according to the logit gap between the gold token and its 
strongest competitor. TrimSFT trims supervision away from both extremes: tokens already mastered (large logit gap) and tokens weakly supported by the current model (small or negative logit gap), concentrating learning within an intermediate logit-gap region between them. We instantiate this principle with a Gaussian weight centered at margin $m$ with bandwidth $\tau$, requiring no reference model or additional forward pass. We evaluate TrimSFT on six base models from the Llama, Qwen, and DeepMath families across five mathematical reasoning benchmarks. TrimSFT consistently improves over standard SFT, achieving the best average performance on five out of six models, with gains of up to $+26.9$ points over SFT on MATH500. Further analyses show that the bandwidth $\tau$ matters more than the exact margin location, and that half-trim variants that remove supervision pressure from only one side yield inferior trade-offs. A token-level logit-gap distribution analysis suggests that TrimSFT reshapes model confidence in a more balanced way than uniform SFT or monotonic reweighting methods. These results suggest that reasoning SFT can benefit from trimming both extremes rather than treating all tokens~uniformly.
\end{abstract}

\section{Introduction}

Large language models (LLMs) have demonstrated strong capabilities on complex reasoning tasks~\citep{wei2022chain, wang2024mmlu, fan2024nphardeval}, with mathematical reasoning serving as a key testbed for studying multi-step inference~\citep{jia2025makes, liu2025safe}. Supervised fine-tuning (SFT) is widely used to adapt pretrained models to mathematical reasoning data~\citep{yu2024metamath, yue2024mammoth}, and often serves as an initialization stage for downstream reinforcement learning or preference optimization~\citep{ouyang2022training,bai2022training}.

Despite its effectiveness, standard SFT applies token-level cross-entropy with uniform weighting across all target tokens~\citep{lin2026sft}, ignoring that different tokens can provide very different learning signals~\citep{wu2026on, gong2026vcore}. In reasoning trajectories, some tokens may already be well mastered and continue to receive unnecessary sharpening pressure, which can contribute to over-confidence~\citep{pereyra2017regularizing, chen2025rethinking, wei2022mitigating}. Other tokens may be highly uncertain, noisy, or beyond the model's current capability, yet still induce large losses and dominate the optimization signal. These two extremes suggest that treating all tokens uniformly can pull learning away from a potentially useful intermediate region~\citep{wu2026on, lin2017focal,han2018co}. This motivates a selective training objective that reduces supervision at both ends of the logit-gap spectrum while concentrating learning on an intermediate region.

\begin{figure}[htbp]
    \centering
    \includegraphics[width=\linewidth]{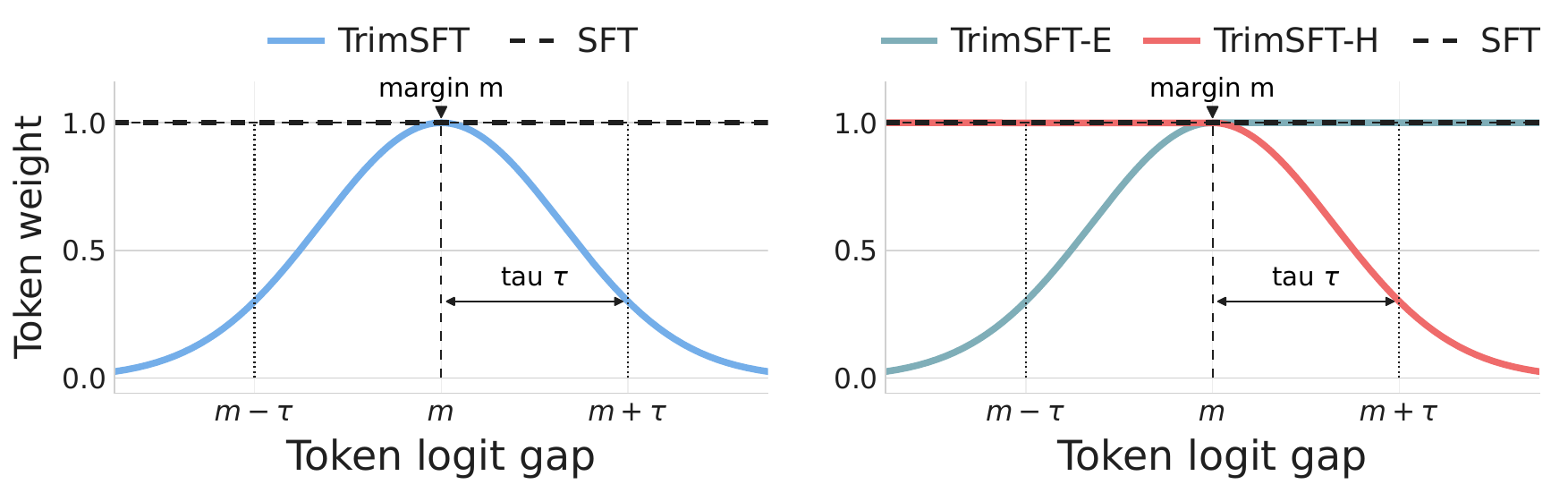}
    \caption{Token weighting functions for SFT, TrimSFT, and its half-trim variants TrimSFT-E and TrimSFT-H. The margin $m$ determines the center of the weighting function, and $\tau$ controls the width of the weighted region.}
    \label{fig:weighting_functions}
\end{figure}

We therefore propose \textbf{Trimmed Logit-Gap SFT (TrimSFT)}, a token-level reweighting method that trims supervision away from both extremes: tokens already mastered (large logit gap) and tokens weakly supported by the current model (small or negative logit gap). For each target token, we compute its \emph{logit gap}, defined as the margin between the gold token's logit and that of its strongest competitor, and use a Gaussian weight centered at margin $m$ with bandwidth $\tau$ to scale the cross-entropy loss (Figure~\ref{fig:weighting_functions}, left). Tokens whose logit gaps lie near $m$ receive stronger supervision, while tokens with much smaller or larger gaps are softly down-weighted, concentrating learning within a bounded intermediate region of the logit-gap spectrum. The weight is computed from the model's own logits in the same forward pass, requiring no reference model or additional forward pass. To examine the contribution of each side of this two-sided trimming profile, we further introduce two half-trim variants (Figure~\ref{fig:weighting_functions}, right): \textbf{Trim-Easy SFT (TrimSFT-E)}, which trims the hard side only and preserves full weight on high-gap (easy) tokens, and \textbf{Trim-Hard SFT (TrimSFT-H)}, which trims the easy side only and preserves full weight on low-gap (hard) tokens.

We evaluate TrimSFT on six base models from the Llama, Qwen, and DeepSeekMath families across five mathematical reasoning benchmarks. Our results show that TrimSFT consistently improves over standard SFT, and we further analyze the mechanism behind these gains. Our contributions are summarized as follows:

\begin{itemize}[leftmargin=1.2em, itemsep=0.25em, topsep=0.25em, parsep=0pt]
    \item We introduce \textbf{TrimSFT}, a token-level reweighting method that scales the SFT loss by the logit gap between the gold token and its strongest competitor. It requires no reference model or additional forward pass.

    \item TrimSFT achieves the best average performance on five out of six models, with gains of up to $+26.9$ points over SFT on MATH500. It also improves capability coverage and self-consistency under repeated sampling, as measured by pass@8 and best-of-8, respectively.

    \item Ablations show that performance is more sensitive to the bandwidth $\tau$ than to the margin $m$, suggesting that the width of the selected logit-gap region is more important than its center.

    \item Half-trim variants show that trimming only one side yields inferior trade-offs: TrimSFT-H collapses below SFT, and TrimSFT-E degrades on the hardest problems. Token-level logit-gap distribution analysis further shows that TrimSFT reshapes model confidence in a more balanced way than uniform SFT or monotonic reweighting methods.
\end{itemize}

\section{Related Work}
\paragraph{Supervised Fine-Tuning for Mathematical Reasoning.}
Mathematical reasoning has emerged as a central testbed for evaluating 
the multi-step inference capabilities of large language 
models~\citep{cobbe2021training, hendrycks2021measuring, 
he2024olympiadbench, saxton2019analysing, lewkowycz2022solving}. A 
common recipe for adapting pretrained models to mathematical tasks is 
supervised fine-tuning (SFT) on curated reasoning 
trajectories~\citep{yu2024metamath, yue2024mammoth, 
toshniwal2024openmathinstruct, li2024numinamath}. Much of the progress 
in this line has come from data-centric improvements, including 
synthesizing high-quality chain-of-thought 
solutions~\citep{yu2024metamath, luo2025wizardmath}, distilling from 
stronger teachers~\citep{shao2024deepseekmath, yang2024qwen2}, 
incorporating recent long chain-of-thought trajectories from 
reasoning-specialized models~\citep{guo2025deepseek, face2025open, 
ye2025limo}, and constructing curated datasets through difficulty- or 
correctness-based filtering~\citep{toshniwal2024openmathinstruct, 
li2024numinamath}. Despite these advances, the underlying objective 
typically remains the standard token-level cross-entropy loss applied 
uniformly to every token. In contrast, our work targets the per-token 
objective rather than the training data, making it complementary to 
existing data-centric approaches.

\paragraph{Token-Level Reweighting and Selection in SFT.}
While vanilla SFT applies a uniform cross-entropy loss across all 
tokens, recent work has explored token-level reweighting or selection 
to account for differences in training value across 
tokens~\citep{lin2024not, ruan2025enhancing, wu2026on}. These methods 
differ mainly in the signals used to score tokens and in the form of 
loss modulation. Some approaches rely on auxiliary signals or 
pre-computed token masks, such as reference-model-based scoring in 
Rho-1~\citep{lin2024not} and counterfactual selection in 
CFT~\citep{ruan2025enhancing}. DFT~\citep{wu2026on} and Focal 
Loss~\citep{lin2017focal} both rescale token-level cross-entropy 
using the predicted probability of the gold token, though in opposite 
monotonic directions: DFT up-weights tokens already assigned high 
probability, while Focal Loss down-weights them. Our method departs 
from these probability-based schemes in two ways: it uses the 
\emph{logit gap} as a more scale-sensitive signal, and applies a 
Gaussian band-pass weighting that targets a bounded intermediate 
region rather than following a monotonic trend. We elaborate on this 
comparison in Section~\ref{sec:unified}.

\paragraph{Logit Gaps, Margins, and Confidence Shaping.}
Margin-related quantities have been widely used to shape confidence 
and compare competing predictions, but they are rarely used directly 
as token-level supervision weights. In classification calibration, 
label smoothing~\citep{muller2019does, pereyra2017regularizing} and 
margin-based label smoothing~\citep{liu2022devil} shape confidence 
and margin behavior to mitigate overconfidence, while logit 
normalization~\citep{wei2022mitigating} and calibration-oriented 
training objectives~\citep{guo2017calibration} regulate confidence 
and logit magnitude during training. Related concerns have also been 
studied in large language models, where recent work examines confidence 
calibration and overconfidence in generated outputs~\citep{zhang2024calibrating, leng2024taming}. Recent evidence further shows that linguistic confidence can diverge substantially from internal, logit-based confidence, highlighting the importance of distinguishing verbalized confidence from model-internal confidence signals~\citep{zhang2026linguistic}. In preference optimization, methods such as SimPO~\citep{meng2024simpo} use sequence-level logit margins between preferred and rejected outputs. These works use margin-related quantities either to regulate confidence or to define sequence-level preference signals. In contrast, TrimSFT uses the per-token logit gap as a supervision weight  during SFT, trimming both extremes of the token-confidence spectrum rather than constraining or maximizing margins globally.

\section{Method}

We begin from the standard supervised fine-tuning~(SFT) objective and then introduce TrimSFT, a token-level reweighting scheme driven by the model's own logit gaps. The key idea is to allocate stronger supervision to tokens whose logits place them near a bounded decision region, while reducing learning pressure on tokens that are already well mastered or currently beyond the model's effective reach. TrimSFT deliberately reduces supervision pressure at both ends of the logit-gap spectrum: tokens with very large gaps (already mastered) receive lower weight, as do tokens with very small or negative gaps (weakly supported by the current model).

Given an input prompt $x$ and a target token sequence $y_1, \dots, y_T$, standard SFT trains a model $\pi_\theta$ by minimizing the token-level cross-entropy loss
\begin{equation}
  \mathcal{L}_{\text{SFT}}
  = -\sum_{t=1}^{T}\log \pi_\theta(y_t \mid x, y_{<t}),
\end{equation}
which assigns uniform weight to every target token. Let $z_t \in \mathbb{R}^{V}$ denote the pre-softmax logits at position $t$, and let $z_{t,y_t}$ denote the logit assigned to the gold token $y_t$. We define the \emph{logit gap} at position $t$ as
\begin{equation}
  \Delta_t = z_{t,y_t} - \max_{v \neq y_t} z_{t,v},
\end{equation}
namely, the margin between the gold token's logit and that of its strongest competing token. A positive $\Delta_t$ indicates that the gold token is currently the top-1 prediction, while a larger magnitude reflects greater separation from the closest alternative. This quantity provides a local, token-level measure of how decisively the model distinguishes the correct token from competing candidates.

\subsection{Trimmed Logit-Gap SFT}

Our goal is to trim supervision away from both extremes of the logit-gap spectrum and concentrate learning on tokens whose gaps fall within an intermediate region. Intuitively, tokens with very large positive gaps are already well separated and may benefit little from continued sharpening, whereas tokens with very small or negative gaps may be highly uncertain, unstable, or beyond the model's current capability. We therefore seek a weighting function that peaks around a moderate logit-gap region and decays on both sides.

We instantiate this with a Gaussian weight on the logit gap:
\begin{equation}
  w_t = \exp\!\left(
    -\frac{(\Delta_t - m)^2}{2\tau^2}
  \right),
\end{equation}
where $m$ is the center of the weighting function and $\tau$ controls the width of the high-weight region. Tokens whose logit gaps lie near $m$ receive the largest weights, while tokens whose gaps are substantially smaller or larger are smoothly down-weighted. In this sense, $m$ determines \emph{where} learning should be concentrated, and $\tau$ determines \emph{how broadly} that concentration should spread.

The resulting TrimSFT objective is
\begin{equation}
\mathcal{L}_{\text{TrimSFT}}
= -\sum_{t=1}^{T} \operatorname{sg}(w_t)\,
\log \pi_\theta(y_t \mid x, y_{<t}),
\label{eq:GB_obj}
\end{equation}
where $\operatorname{sg}(\cdot)$ denotes stop-gradient. The weights $w_t$ are computed from the same forward pass as the logits and then detached, so they act purely as a rescaling per-token of the cross-entropy loss without contributing a gradient through $\theta$. This does not require any auxiliary model or extra forward pass, and adds only minimal computational overhead.

Compared with probability-based reweighting, the logit gap is particularly suitable in our setting for two reasons. First, it directly measures the separation between the gold token and its strongest competitor, making it naturally aligned with how close a token is to the model's decision boundary. Second, unlike the gold-token probability, which tends to saturate in high-confidence regimes, the logit gap remains discriminative even when probabilities would otherwise appear nearly indistinguishable. The Gaussian weighting then realizes a smooth two-sided trimming profile over token states: supervision is concentrated on tokens whose logit gaps lie near the chosen margin $m$, and decays on both sides. Unlike monotonic reweighting schemes that progressively emphasize either harder or easier tokens across the full confidence range, TrimSFT explicitly trims both extremes and focuses learning on a bounded intermediate region.

\subsection{Half-Trim Variants}
\label{sec:variants}

The weighting function in TrimSFT is symmetric around the margin $m$, assigning lower weights to tokens on both sides as their logit gaps move away from $m$. To isolate the contribution of each side, we define two half-trim variants that retain Gaussian decay on only one side while keeping full weight on the other (Figure~\ref{fig:weighting_functions}, right).

\paragraph{TrimSFT-E.}
TrimSFT-E trims the hard side only: it keeps full weight on tokens whose logit gaps exceed the margin $m$, and applies Gaussian decay only on the lower-gap side:
\begin{equation}
  w_t^{\text{E}} =
  \begin{cases}
    \exp\!\left(-\dfrac{(\Delta_t - m)^2}{2\tau^2}\right), & \Delta_t < m, \\[6pt]
    1, & \Delta_t \geq m.
  \end{cases}
\end{equation}
This preserves full supervision for tokens already beyond the chosen margin, while down-weighting tokens with smaller gaps.

\paragraph{TrimSFT-H.}
TrimSFT-H trims the easy side only: it keeps full weight on tokens whose logit gaps fall below the margin $m$, and applies Gaussian decay only on the higher-gap side:
\begin{equation}
  w_t^{\text{H}} =
  \begin{cases}
    1, & \Delta_t \leq m, \\[6pt]
    \exp\!\left(-\dfrac{(\Delta_t - m)^2}{2\tau^2}\right), & \Delta_t > m.
  \end{cases}
\end{equation}
This preserves full supervision for tokens with smaller gaps, while down-weighting tokens that are already well separated.

Together, these variants retain the same margin $m$ and bandwidth $\tau$ as TrimSFT, but isolate the contribution of each side of the weighting profile. Comparing them with full TrimSFT allows us to determine whether its gains arise primarily from trimming high-gap tokens, low-gap tokens, or both.

\subsection{A Unified View of Token Reweighting}
\label{sec:unified}

Standard SFT, TrimSFT, and several representative reweighting methods, including DFT~\citep{wu2026on} and a focal-loss-based SFT baseline (FSFT)~\citep{lin2017focal}, can all be written in a unified token-reweighted form:
\begin{equation}
  \mathcal{L} = \sum_{t=1}^{T} \operatorname{sg}(w_t)\,\ell_t,
\end{equation}
where $\ell_t = -\log \pi_\theta(y_t \mid x, y_{<t})$ is the standard token-level cross-entropy loss, $\operatorname{sg}(\cdot)$ denotes stop-gradient, and the methods differ only in the choice of $w_t$. Representative objectives then correspond to different weighting functions:
\begin{equation}
w_t =
\begin{cases}
1, & \text{SFT}, \\[4pt]
(1 - p_{t,y_t})^\gamma, & \text{FSFT}, \\[4pt]
p_{t,y_t}, & \text{DFT}, \\[4pt]
\exp\!\left(-\dfrac{(\Delta_t - m)^2}{2\tau^2}\right), & \text{TrimSFT}.
\end{cases}
\end{equation}

In this framework, standard SFT assigns equal training weight to every token. FSFT and DFT instead reshape the SFT objective using the gold-token probability: FSFT places larger weights on lower-confidence tokens, thereby emphasizing harder or less well-mastered positions, whereas DFT increases the relative contribution of higher-probability tokens to compensate for the implicit inverse-probability bias of standard cross-entropy. In contrast, TrimSFT uses the logit gap $\Delta_t$ and applies a non-monotonic weighting profile that trims both extremes while concentrating learning on tokens whose gaps lie near a prescribed margin $m$. 

This unified view makes two differences especially clear. First, TrimSFT operates on the logit gap rather than the probability, so it directly tracks the separation between the gold token and its strongest competitor and avoids the saturation that compresses high-confidence tokens to nearly identical weights under probability-based schemes. Second, TrimSFT targets a bounded intermediate region between well-mastered and weakly supported tokens, whereas existing probability-based reweighting methods follow a single monotonic trend over the full confidence range.

\section{Experiments}

\subsection{Setup}
\paragraph{Dataset.} 
\texttt{NuminaMath-CoT}~\citep{li2024numinamath} is a large-scale math reasoning dataset containing about 860K problem-solution pairs with chain-of-thought annotations. For all experiments, we randomly sample 20,000 problems for supervised fine-tuning.

\paragraph{Models.} 
We evaluate TrimSFT and the baseline methods on six base models from three families, spanning parameter scales from 1.5B to 8B: Llama3.2-3B~\citep{meta2024llama32}, Llama3.1-8B~\citep{grattafiori2024llama}, DeepSeekMath-7B~\citep{shao2024deepseekmath}, Qwen2.5-Math-1.5B, Qwen2.5-Math-7B~\citep{yang2024qwen2}, and Qwen3-4B-Base~\citep{yang2025qwen3}. We use only base models, rather than instruction-tuned variants, to reduce the influence of prior instruction tuning and enable a cleaner comparison of different supervised fine-tuning objectives.

\paragraph{Baselines.}
We compare \textbf{TrimSFT} with four representative baselines: (i) \textbf{Base}, the original pretrained model without supervised fine-tuning; (ii) \textbf{SFT}, standard supervised fine-tuning with uniform token-level weighting; (iii) \textbf{FSFT}, a focal-loss-based variant that upweights lower-confidence tokens during training~\citep{lin2017focal}; and (iv) \textbf{DFT}~\citep{wu2026on}, a probability-based reweighting approach that assigns larger weights to higher-probability tokens.

\paragraph{Evaluation benchmarks.} 
We evaluate all methods on five mathematical reasoning benchmarks spanning
a range of difficulty levels: MATH500~\citep{hendrycks2021measuring},
OlympiadBench~\citep{he2024olympiadbench},
Minerva~\citep{lewkowycz2022solving}, AMC~\citep{aimo2024amc}, and
AIME24~\citep{aimo2024aime}. For each benchmark, we sample \(N=8\)
generations per problem and report \textbf{average@8} as the primary
metric, which reflects the model's average generation quality. In
Section~\ref{sec:passk}, we additionally report \textbf{pass@8} and
\textbf{best-of-8}, where majority voting is used as the selector, to
characterize the model's exploration ability and self-consistency,
respectively. Detailed evaluation settings are provided in
Appendix~\ref{app:exp-setting}.

\paragraph{Implementation details.} 
For TrimSFT, we use a single hyperparameter setting, 
\((m,\tau)=(1.5,0.8)\), for all models and benchmarks in the main 
comparison (Table~\ref{tab:main}), without any model- or 
benchmark-specific tuning. All results are reported using the 
checkpoint obtained after one training epoch. We study the effects of 
varying \(m\) and \(\tau\) in Sections~\ref{subsec:ablation} 
and~\ref{Asym-variants}, and provide additional training and evaluation 
details in Appendix~\ref{app:exp-setting}. Code is available at \url{https://github.com/karpning/TrimSFT}.

\subsection{Main Results}
\label{main-result}
Table~\ref{tab:main} reports the main results under the \textbf{average@8} accuracy across six base models and five mathematical reasoning benchmarks, with TrimSFT evaluated using $(m,\tau)=(1.5, 0.8)$.

\begin{table*}[t]
\centering
\scriptsize
\setlength{\tabcolsep}{3.5pt}
\caption{Average@8 accuracy (\%) on math reasoning benchmarks. \textbf{Bold} marks the best result and \underline{underlined} marks the second-best within each model block. Avg. denotes the mean across benchmarks.}
\label{tab:main}
\resizebox{0.9\textwidth}{!}{
\begin{tabular}{l|l|ccccc|c}
\toprule
\textbf{Model} & \textbf{Method} & \textbf{MATH500} & \textbf{OlyBench} & \textbf{Minerva} & \textbf{AMC} & \textbf{AIME24} & \textbf{Avg.} \\
\midrule
\multirow{5}{*}{Llama3.2-3B}
 & Base    & 1.50  & 0.86  & 0.66  & \underline{1.48}  & 0.00 & 0.90 \\
 & SFT     & 4.80  & 1.56  & 1.65  & \underline{1.48}  & 0.00 & 1.90 \\
 & FSFT    & 1.90  & 0.96  & 0.83  & 1.19  & 0.00 & 0.98 \\
 & DFT     & \underline{8.78}  & \underline{2.35} & \textbf{3.60} & 1.05 & 0.00 & \underline{3.16} \\
 & \textbf{TrimSFT} & \textbf{9.45}  & \textbf{2.81} & \underline{2.92} & \textbf{3.53} & \textbf{2.00} & \textbf{4.14} \\
\midrule
\multirow{5}{*}{Llama3.1-8B}
 & Base    & 2.10  & 0.93  & 1.65  & 1.20  & 0.00 & 1.18 \\
 & SFT     & 12.05 & 2.59  & 3.26  & 3.40  & 0.41 & 4.34 \\
 & FSFT    & 2.95  & 0.89  & 0.96  & 1.64  & 0.00 & 1.29 \\
 & DFT     & \underline{19.77} & \underline{4.89} & \textbf{5.81} & \underline{4.73} & \underline{0.83} & \underline{7.21} \\
 & \textbf{TrimSFT} & \textbf{20.95} & \textbf{5.13} & \underline{5.52} & \textbf{6.17} & \textbf{1.41} & \textbf{7.84} \\
\midrule
\multirow{5}{*}{DeepSeekMath-7B}
 & Base    & 5.45  & 1.63  & 2.15  & 1.94  & 0.00 & 2.23 \\
 & SFT     & 23.40 & 5.52  & 7.36  & 7.08  & 0.41 & 8.75 \\
 & FSFT    & 6.80  & 1.70  & 2.26  & 2.80  & 0.00 & 2.71 \\
 & DFT     & \underline{38.42} & \underline{12.16} & \textbf{13.30} & \textbf{13.23} & \textbf{1.24} & \textbf{15.67} \\
 & \textbf{TrimSFT} & \textbf{38.80} & \textbf{12.94} & \underline{12.26} & \underline{12.35} & \textbf{1.24} & \underline{15.52} \\
\midrule
\multirow{5}{*}{Qwen2.5-Math-1.5B}
 & Base    & 31.45 & 16.23 & 8.45  & 13.68 & 3.32 & 14.63 \\
 & SFT     & 40.02 & 12.11 & 10.08 & 12.64 & 2.50 & 15.47 \\
 & FSFT    & 10.30 & 1.93  & 2.34  & 1.76  & 0.00 & 3.27 \\
 & DFT     & \underline{63.17} & \underline{26.90} & \underline{20.55} & \textbf{27.95} & \underline{7.93} & \underline{29.30} \\
 & \textbf{TrimSFT} & \textbf{66.95} & \textbf{29.41} & \textbf{24.96} & \textbf{27.95} & \textbf{9.58} & \textbf{31.77} \\
\midrule
\multirow{5}{*}{Qwen2.5-Math-7B}
 & Base    & 40.07 & 18.45 & 13.43 & 18.50 & \textbf{11.66} & 20.42 \\
 & SFT     & 50.65 & 16.91 & 17.46 & 18.09 & 1.66 & 20.95 \\
 & FSFT    & 18.12 & 3.06 & 4.02 & 4.12 & 0.41 & 5.95 \\
 & DFT     & \underline{69.75} & \underline{33.08} & \underline{26.31} & \underline{34.41} & \textbf{11.66} & \underline{35.04} \\
 & \textbf{TrimSFT} & \textbf{71.15} & \textbf{33.61} & \textbf{28.88} & \textbf{37.48} & 8.76 & \textbf{35.98} \\
\midrule
\multirow{5}{*}{Qwen3-4B-Base}
 & Base    & 34.35 & 17.96 & 11.76 & 17.06 & \underline{4.15} & 17.06 \\
 & SFT     & 47.45 & 15.90 & 15.10 & 16.16 & 1.66 & 19.25 \\
 & FSFT    & 12.40 & 1.97 & 3.16 & 3.39 & 0.41 & 4.27 \\
 & DFT     & \underline{61.65} & \textbf{27.37} & \underline{22.88} & \underline{24.74} & \textbf{5.41} & \underline{28.41} \\
 & \textbf{TrimSFT} & \textbf{61.80} & \underline{26.08} & \textbf{25.03} & \textbf{27.93} & \underline{4.15} & \textbf{29.00} \\
\bottomrule
\end{tabular}
}
\end{table*}

Overall, TrimSFT consistently improves over standard SFT across all six models and achieves the best average performance on five out of six models. The gains are especially large on math-oriented models: for example, TrimSFT improves the average score from $15.47$ to $31.77$ on Qwen2.5-Math-1.5B and from $20.95$ to $35.98$ on Qwen2.5-Math-7B. On MATH500, TrimSFT is the top-performing method across all six models, with the largest gain over SFT reaching $+26.93$ points on Qwen2.5-Math-1.5B ($40.02 \rightarrow 66.95$). Compared with DFT, TrimSFT achieves stronger overall average performance on five of the six models, suggesting that logit-gap-based reweighting provides a useful alternative to probability-based token weighting. In contrast, FSFT performs poorly in most settings, indicating that simply emphasizing harder tokens is insufficient for reasoning SFT. These results provide initial evidence for our hypothesis that effective supervised fine-tuning for mathematical reasoning benefits from trimming both extremes of the logit-gap spectrum rather than treating all tokens~uniformly.

\subsection{Capability Ceiling and Self-Consistency}
\label{sec:passk}

Beyond \textbf{average@8}, we further evaluate \textbf{pass@8} and
\textbf{best-of-8} with majority voting to characterize two complementary
aspects of model behavior. Pass@8 measures whether the model can produce
at least one correct solution among multiple samples, reflecting its
capability ceiling or exploration coverage. Best-of-8 with majority
voting measures whether the model's sampled solutions consistently
support the correct answer, reflecting the self-consistency of its
generation distribution.

\begin{figure*}[htbp]
    \centering
    \includegraphics[width=0.95\textwidth]{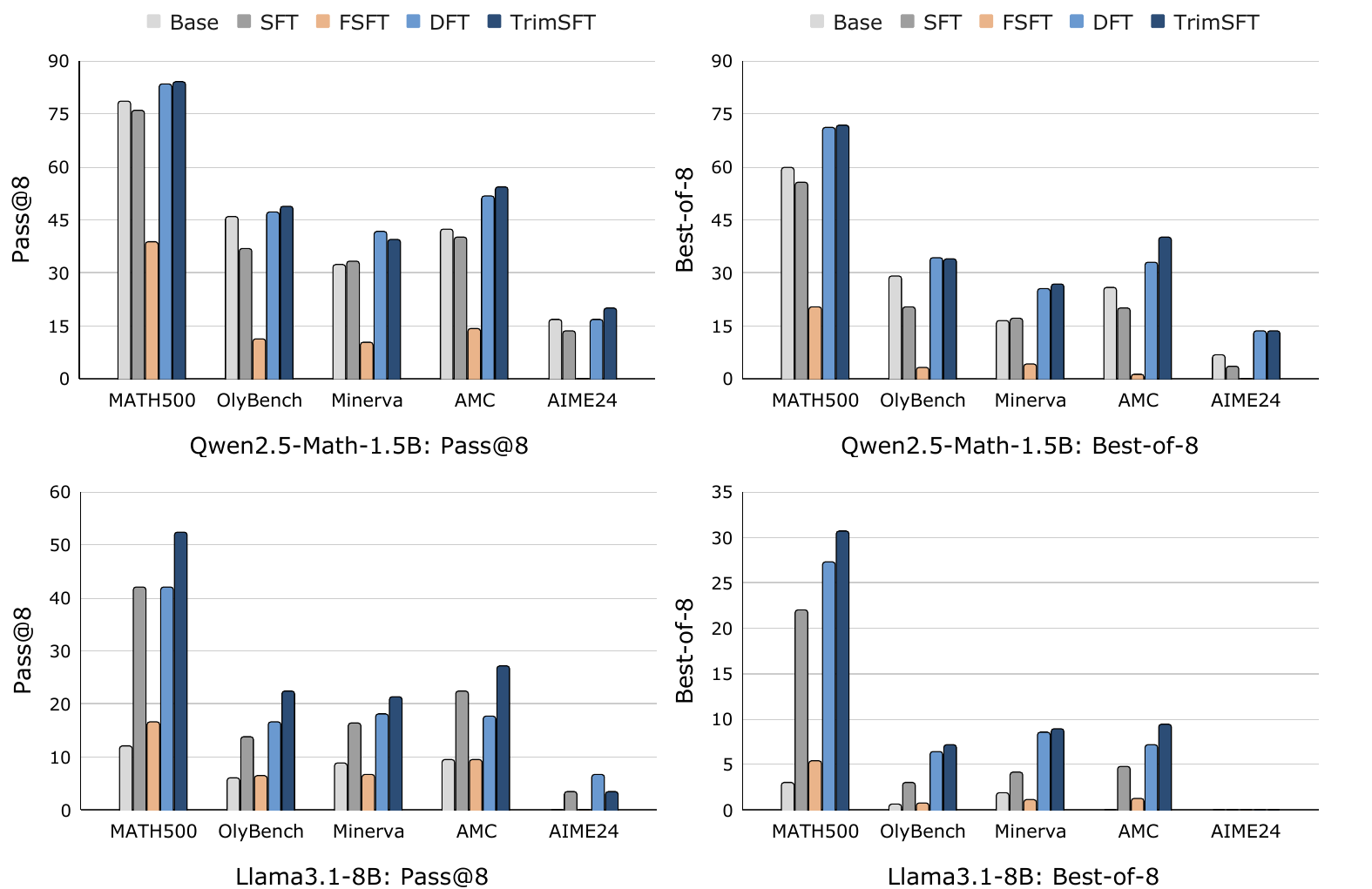}
    \caption{Pass@8 and best-of-8 with majority voting on Qwen2.5-Math-1.5B and Llama3.1-8B across five mathematical reasoning benchmarks.}
    \label{fig:passk_results}
\end{figure*}

We report results on Qwen2.5-Math-1.5B and Llama3.1-8B in
Figure~\ref{fig:passk_results}. For both models, TrimSFT uses the same
fixed hyperparameter setting \((m,\tau)=(1.5,0.8)\) as in the main
comparison in Table~\ref{tab:main}.

As shown in Figure~\ref{fig:passk_results}, TrimSFT consistently improves
pass@8 over standard SFT on both models across most benchmarks, indicating
that two-sided logit-gap trimming expands the model's ability to discover
correct solutions under repeated sampling. This suggests that TrimSFT does
not merely optimize a single deterministic output, but improves the broader
solution space explored by the model. The improvement also extends to
best-of-8 with majority voting, where TrimSFT remains stronger than or
competitive with the baselines across the two models. Since majority voting
requires multiple samples to converge toward the correct answer, these gains
indicate that the sampled solutions become more reliably aligned rather than
only occasionally correct. Together with the average@8 results in
Table~\ref{tab:main}, these findings show that TrimSFT improves average
generation quality, capability coverage, and self-consistency under repeated
sampling.

\subsection{Ablation: Margin and Bandwidth}
\label{subsec:ablation}

\begin{figure}[htbp]
    \centering
    \includegraphics[width=\linewidth]{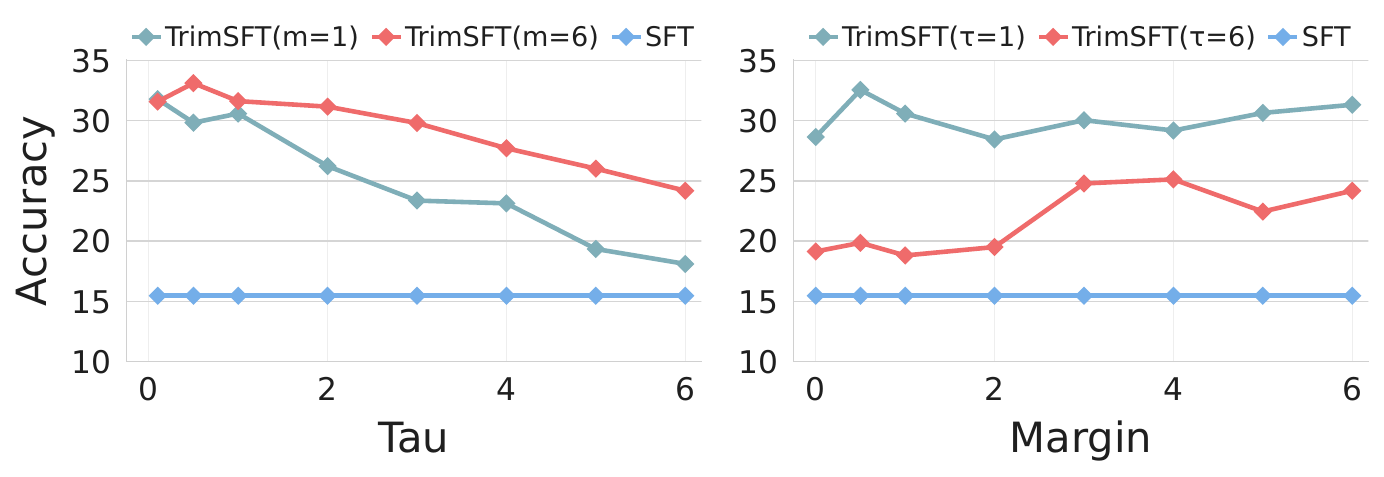}
    \caption{Ablation of margin $m$ and bandwidth $\tau$ on Qwen2.5-Math-1.5B. Left: varying $\tau$ with fixed $m\in\{1,6\}$; right: varying $m$ with fixed $\tau\in\{1,6\}$. The dashed line denotes the average SFT baseline.}
    \label{fig:ablation}
\end{figure}

We further study the effect of the two key hyperparameters in TrimSFT: the margin $m$, which determines the center of the weighted logit-gap region, and the bandwidth $\tau$, which controls how broadly tokens around this margin are emphasized. Figure~\ref{fig:ablation} reports average@8 accuracy, averaged across the five benchmarks, on Qwen2.5-Math-1.5B under different choices of $m$ and $\tau$, with the full numerical results provided in Appendix~\ref{app:ablation}. 

Figure~\ref{fig:ablation} shows that TrimSFT is more sensitive to the bandwidth $\tau$ than to the margin $m$. When $m$ is fixed, increasing $\tau$ leads to a clear performance drop, especially for $m=1$, indicating that an overly broad weighting function weakens the intended focus on an intermediate logit-gap region. In contrast, when $\tau=1$ is fixed, TrimSFT remains consistently strong across different values of $m$ and stays well above the SFT baseline, with only moderate variation as the margin changes. This relative insensitivity to $m$ holds primarily when the bandwidth is small: with $\tau=6$, performance varies more visibly with $m$, whereas with $\tau=1$ the curve remains comparatively stable. These results suggest that the width of the selected region is more important than its exact center: TrimSFT benefits most when supervision is concentrated within a relatively narrow band of logit gaps, under which the choice of $m$ becomes less critical. Additional token-weight diagnostics further show that $\tau$ primarily controls the selectivity and trimming strength of the objective, while $m$ mainly shifts the selected logit-gap region; detailed results are provided in Appendix~\ref{role-margin-tau}.

\subsection{Half-Trim Variants}
\label{Asym-variants}
We compare TrimSFT with its two half-trim variants, TrimSFT-E and TrimSFT-H, to examine whether the full two-sided trimming profile is necessary. TrimSFT-E trims the hard side only and keeps easy tokens at full weight, whereas TrimSFT-H trims the easy side only and keeps hard tokens at full weight. To avoid drawing conclusions from a single margin choice, we evaluate the three variants under representative margins $m\in\{1,3,5\}$ while fixing $\tau=1$. Table~\ref{tab:variants} reports results on three representative benchmarks.

\begin{table}[htbp]
\centering
\small
\setlength{\tabcolsep}{5pt}
\caption{Comparison between TrimSFT and its half-trim variants on Qwen2.5-Math-1.5B with fixed $\tau=1$ across representative margins $m\in\{1,3,5\}$. \textbf{Bold} marks the best result among the three variants.}
\label{tab:variants}
\begin{tabular}{llccc}
\toprule
\textbf{$m$} & \textbf{Method} & \textbf{MATH500} & \textbf{AMC} & \textbf{AIME24} \\
\midrule
\multirow{3}{*}{1}
& TrimSFT  & 62.80 & 31.74 & \textbf{12.07} \\
& TrimSFT-E & \textbf{65.98} & \textbf{33.01} & 7.51 \\
& TrimSFT-H & 28.75 & 8.84 & 0.41 \\
\midrule

\multirow{3}{*}{3}
& TrimSFT  & \textbf{65.98} & \textbf{30.29} & \textbf{7.50} \\
& TrimSFT-E & 64.30 & 26.18 & 4.56 \\
& TrimSFT-H & 40.92 & 13.39 & 2.06 \\
\midrule

\multirow{3}{*}{5}
& TrimSFT  & 65.96 & 31.03 & \textbf{7.50} \\
& TrimSFT-E & \textbf{66.67} & \textbf{32.19} & 6.25 \\
& TrimSFT-H & 42.50 & 16.03 & 1.65 \\
\bottomrule
\end{tabular}
\end{table}

Table~\ref{tab:variants} shows a consistent asymmetry between the two half-trim variants. TrimSFT-E is often competitive with or slightly stronger than TrimSFT on MATH500 and AMC, suggesting that preserving full supervision for higher-gap tokens can benefit relatively easier or medium-difficulty benchmarks. However, this advantage does not consistently transfer to the harder AIME24 benchmark, where TrimSFT achieves the best performance across all three reported margins. In contrast, TrimSFT-H performs substantially worse across all settings,
mirroring the poor performance of FSFT in Table~\ref{tab:main}. These
results reveal an asymmetric contribution from the two sides of the
weighting profile: suppressing low-gap tokens appears to be the primary
source of stability, whereas trimming high-gap tokens provides an
additional regularization effect. This helps explain why TrimSFT-E can
remain competitive on easier benchmarks while full TrimSFT provides a
more robust trade-off on harder problems. Additional gradient-mass
analysis supporting this interpretation is provided in
Appendix~\ref{app:gradient-mass}.

\section{Mechanism: Logit Gap Distribution Analysis}

\begin{figure}[htbp]
    \centering
    \includegraphics[width=\linewidth]{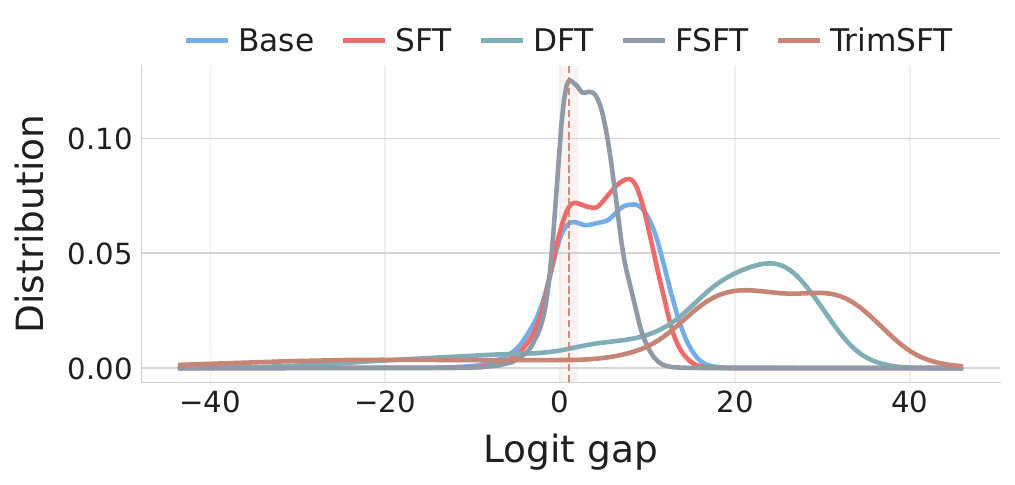}
    \caption{Logit-gap distributions on 100 randomly sampled training examples for Qwen2.5-Math-1.5B, aggregated over all response tokens.}
    \label{fig:gap_distribution}
\end{figure}

To better understand how different fine-tuning objectives reshape token-level confidence, i.e., the model's margin-based confidence in the gold response tokens, we analyze the distribution of logit gaps on the training data. We randomly sample 100 examples from the training set and compute teacher-forced logit gaps for every token in the response part of each example. We then aggregate all token-level gaps across the sampled examples and plot their kernel density estimates for Base, SFT, DFT, FSFT, and TrimSFT in Figure~\ref{fig:gap_distribution}.

The resulting distributions reveal clear differences in how each objective reshapes token-level confidence. Standard SFT shifts the distribution to the right relative to the base model, indicating stronger separation between gold tokens and their competitors after fine-tuning. FSFT, in contrast, leaves more mass near small logit gaps, consistent with its emphasis on low-confidence tokens and its poor performance in Table~\ref{tab:main}. DFT also moves the distribution toward larger gaps and appears closer to TrimSFT than to standard SFT, which may help explain why DFT often achieves competitive performance. However, DFT and TrimSFT reach this behavior through different mechanisms: DFT monotonically emphasizes high-probability tokens, whereas TrimSFT trims supervision away from both extremes of the logit-gap spectrum. As a result, TrimSFT shifts the distribution toward larger gaps while preserving a broad shape and reducing mass in the small-gap region. This supports our interpretation that TrimSFT improves reasoning SFT by reshaping token-level confidence in a more balanced way, rather than simply over-emphasizing uncertain tokens or uniformly sharpening already confident ones.

Beyond these distributional analyses, additional token-category results in Appendix~\ref{app:token-category} further characterize how TrimSFT redistributes the optimization signal, showing a larger share of gradient mass on numeric and mathematical-symbol tokens and a reduced contribution from generic other tokens.

\section{Conclusion}

We presented TrimSFT, a token-level reweighting method that uses logit gaps to concentrate supervision on an intermediate confidence region. By trimming both already well-separated tokens and tokens with weak current support, TrimSFT provides a non-monotonic alternative to uniform SFT and monotonic reweighting methods. Across six base models and five mathematical reasoning benchmarks, TrimSFT consistently improves over standard SFT, while further analyses show gains in capability coverage, self-consistency, and better trade-offs than half-trim variants. Mechanistically, TrimSFT reshapes token-level confidence by moving tokens toward more confident regions while avoiding uniform over-sharpening. Overall, our results suggest that reasoning SFT can benefit from
selectively concentrating supervision within an intermediate logit-gap
region rather than applying uniform pressure across all tokens.

\section*{Limitations}

TrimSFT is evaluated primarily on mathematical reasoning benchmarks with supervised fine-tuning from base models, so its effectiveness on broader domains such as code generation, open-ended instruction following, or long-form reasoning remains to be further studied. In addition, TrimSFT uses two hyperparameters, the margin $m$ and bandwidth $\tau$, to define the emphasized logit-gap region. Our ablations show that the method is relatively robust to the exact margin choice and generally improves over standard SFT across a range of settings, suggesting that careful tuning is not necessary to obtain gains. Still, different model scales or data distributions may benefit from adaptive strategies that adjust the weighting region automatically during training.

\section*{Ethical Considerations}

TrimSFT is a token-level reweighting method for supervised fine-tuning and does not involve new data collection or human annotation. All experiments are conducted using publicly available pretrained models and mathematical reasoning datasets intended for research use. As with other supervised fine-tuning methods, improving reasoning capabilities may also strengthen broader generation abilities of language models, which could potentially be misused in downstream applications.

\section*{Acknowledgments}

This research was supported in part by the National Science Foundation under Grant No. 2452367.

\bibliography{ref}
\newpage
\appendix

\startcontents[appendices]
\section*{Contents of Appendix}
\printcontents[appendices]{}{1}{\normalsize}

\section{Experiment Setting}
\label{app:exp-setting}

\paragraph{Parameter setting.}
We train all models for one epoch using the AdamW optimizer with a learning rate of \(5\times10^{-5}\), cosine learning rate decay with warmup, and gradient clipping. Training is conducted with bfloat16 mixed precision and FlashAttention-2 acceleration under the FSDP framework. Unless otherwise specified, the global batch size is 24 with a per-device micro-batch size of 4, and the maximum sequence length is set to 2048. For TrimSFT, we set the margin to \(m=1.5\) and the bandwidth to \(\tau=0.8\).

\paragraph{Baseline setting.}
All four methods, SFT, FSFT, DFT, and TrimSFT, are trained under an
identical configuration, including optimizer, learning rate, schedule,
batch size, sequence length, precision, and number of epochs. They
differ only in the per-token weight $w_t$ applied to the cross-entropy
loss, as summarized in Section~\ref{sec:unified}. The only
method-specific hyperparameters beyond this shared recipe are the
margin $m$ and bandwidth $\tau$ in TrimSFT, and the focusing parameter
$\gamma$ in FSFT, for which we use $\gamma=2$ following common practice
in the focal-loss literature~\citep{lin2017focal}. DFT introduces no
additional hyperparameters.

\paragraph{Evaluation protocol.}
For evaluation, we sample eight generations per problem using temperature
\(1.0\), top-\(p\) sampling with \(p=1.0\), and a maximum generation
length of 2048 tokens. We use temperature \(1.0\) to encourage diverse
sampled solutions, which is necessary for evaluating average@8, pass@8,
and best-of-8 under repeated sampling. Average@8 is computed as the mean
correctness over the eight sampled generations, pass@8 checks whether at
least one generation is correct, and best-of-8 uses majority voting over
the extracted final answers. We extract final answers from the generated
solutions and judge correctness using the same evaluation protocol across
all methods and benchmarks.

\paragraph{Computing infrastructure.}
Experiments are conducted on NVIDIA RTX A6000 GPUs, each with 48GB memory, using CUDA 12.8 and NVIDIA driver 570.207. Each training run uses two GPUs with FSDP unless otherwise specified.

\section{Detailed Results for Ablations}
\label{app:ablation}

Table~\ref{tab:ablation-fixed-m} and Table~\ref{tab:ablation-fixed-tau}
report the full per-benchmark results for the margin and bandwidth ablation study shown in Figure~\ref{fig:ablation}. Each setting is denoted by \((m,\tau)\), where \(m\) is the margin and \(\tau\) is the bandwidth. While the main text summarizes the averaged trends across the evaluated benchmarks, this appendix provides the complete
per-benchmark results for both varying the bandwidth under fixed
margins and varying the margin under fixed bandwidths.

\begin{table}[htbp]
\centering
\scriptsize
\setlength{\tabcolsep}{4pt}
\caption{Ablation results with fixed \(m\) and varying \(\tau\). Each setting is denoted as \((m,\tau)\).}
\label{tab:ablation-fixed-m}
\resizebox{0.5\textwidth}{!}{
\begin{tabular}{lcccccc}
\toprule
\textbf{\((m,\tau)\)} & \textbf{MATH500} & \textbf{OlyBench} & \textbf{Minerva} & \textbf{AMC} & \textbf{AIME24} & \textbf{Avg.} \\
\midrule
\multicolumn{7}{c}{\textit{Fixed \(m=1\)}} \\
\midrule
\((1,0.1)\) & 66.83 & 29.53 & 23.86 & 29.02 & 9.61 & 31.77 \\
\((1,0.5)\) & 63.72 & 27.17 & 23.16 & 25.90 & 9.16 & 29.82 \\
\((1,1)\)   & 62.80 & 27.66 & 18.65 & 31.74 & 12.07 & 30.58 \\
\((1,2)\)   & 60.38 & 24.68 & 18.46 & 25.00 & 4.58 & 26.22 \\
\((1,3)\)   & 53.63 & 21.05 & 16.30 & 22.50 & 3.32 & 23.36 \\
\((1,4)\)   & 54.05 & 21.03 & 16.88 & 20.59 & 3.12 & 23.13 \\
\((1,5)\)   & 46.83 & 16.38 & 14.15 & 17.34 & 2.07 & 19.35 \\
\((1,6)\)   & 46.93 & 15.30 & 12.96 & 16.78 & 2.07 & 18.81 \\
\midrule
\multicolumn{7}{c}{\textit{Fixed \(m=6\)}} \\
\midrule
\((6,0.1)\) & 65.12 & 30.76 & 23.31 & 30.75 & 7.93 & 31.57 \\
\((6,0.5)\) & 68.60 & 30.69 & 26.83 & 31.45 & 7.94 & 33.10 \\
\((6,1)\)   & 66.83 & 29.28 & 22.65 & 27.65 & 11.65 & 31.61 \\
\((6,2)\)   & 66.39 & 28.60 & 21.11 & 33.39 & 8.76 & 31.65 \\
\((6,3)\)   & 63.95 & 27.44 & 20.50 & 29.55 & 7.50 & 29.79 \\
\((6,4)\)   & 60.70 & 26.40 & 19.21 & 26.77 & 5.42 & 27.70 \\
\((6,5)\)   & 56.90 & 24.88 & 18.65 & 25.45 & 4.15 & 26.01 \\
\((6,6)\)   & 54.40 & 22.26 & 16.86 & 22.81 & 4.58 & 24.18 \\
\midrule
  SFT          & 40.02 & 12.11 & 10.08 & 12.64 & 2.50 & 15.47 \\
\bottomrule
\end{tabular}
}
\end{table}

\begin{table}[t]
\centering
\scriptsize
\setlength{\tabcolsep}{4pt}
\caption{Ablation results with fixed \(\tau\) and varying \(m\). Each setting is denoted as \((m,\tau)\).}
\label{tab:ablation-fixed-tau}
\resizebox{0.5\textwidth}{!}{
\begin{tabular}{lcccccc}
\toprule
\textbf{\((m,\tau)\)} & \textbf{MATH500} & \textbf{OlyBench} & \textbf{Minerva} & \textbf{AMC} & \textbf{AIME24} & \textbf{Avg.} \\
\midrule
\multicolumn{7}{c}{\textit{Fixed \(\tau=1\)}} \\
\midrule
\((0,1)\)   & 61.10 & 26.64 & 20.96 & 26.60 & 7.91 & 28.64 \\
\((0.5,1)\) & 66.75 & 30.73 & 23.45 & 31.76 & 9.99 & 32.54 \\
\((1,1)\)   & 62.80 & 27.66 & 18.65 & 31.74 & 12.07 & 30.58 \\
\((2,1)\)   & 62.25 & 26.06 & 19.39 & 27.36 & 7.09 & 28.43 \\
\((3,1)\)   & 65.98 & 26.76 & 19.64 & 30.29 & 7.50 & 30.03 \\
\((4,1)\)   & 63.60 & 26.90 & 22.03 & 27.51 & 5.83 & 29.17 \\
\((5,1)\)   & 65.96 & 28.55 & 20.13 & 31.03 & 7.50 & 30.63 \\
\((6,1)\)   & 66.83 & 29.27 & 22.65 & 27.65 & 11.65 & 31.61 \\
\midrule
\multicolumn{7}{c}{\textit{Fixed \(\tau=6\)}} \\
\midrule
\((0,6)\)   & 46.97 & 15.98 & 13.83 & 16.46 & 2.47 & 19.14 \\
\((0.5,6)\) & 48.13 & 17.14 & 14.23 & 16.03 & 3.75 & 19.86 \\
\((1,6)\)   & 46.93 & 15.30 & 12.96 & 16.78 & 2.07 & 18.81 \\
\((2,6)\)   & 46.30 & 17.74 & 13.88 & 16.32 & 3.32 & 19.51 \\
\((3,6)\)   & 54.58 & 21.48 & 17.74 & 24.26 & 5.84 & 24.78 \\
\((4,6)\)   & 55.88 & 22.75 & 17.19 & 23.96 & 5.82 & 25.12 \\
\((5,6)\)   & 51.50 & 20.25 & 16.53 & 21.46 & 2.49 & 22.45 \\
\((6,6)\)   & 54.40 & 22.26 & 16.86 & 22.81 & 4.57 & 24.18 \\
\midrule
SFT          & 40.02 & 12.11 & 10.08 & 12.64 & 2.50 & 15.47 \\
\bottomrule
\end{tabular}
}
\end{table}

The per-benchmark results are consistent with the averaged trends in
the main text. When the margin \(m\) is fixed, increasing the bandwidth
\(\tau\) generally reduces performance, especially when \(\tau\) becomes
large. In contrast, when \(\tau=1\) is fixed, the method remains stronger
than standard SFT across a wide range of margin values, indicating that
it does not require careful tuning of the exact margin location to
obtain improvements.

\section{Understanding the Roles of $m$ and $\tau$}
\label{role-margin-tau}

To further explain the different sensitivities of TrimSFT to the bandwidth $\tau$ and the margin $m$, we conduct a token-weight analysis under the same experimental setting as Section~\ref{subsec:ablation}, using Qwen2.5-Math-1.5B. Specifically, we compute logit gaps on 40,082 gold response tokens sampled from the training data and derive the corresponding TrimSFT weights:
\begin{equation}
w_t = \exp\!\left(-\frac{(\Delta_t-m)^2}{2\tau^2}\right),
\end{equation}
where $\Delta_t = z_{t,y_t} - \max_{v\neq y_t} z_{t,v}$.

We report two diagnostics. \textbf{High-Weight (\%)} is the fraction of tokens with $w_t>0.5$, while the \textbf{Effective Token Ratio}, denoted as $R_{\mathrm{eff}}$, is defined as
\begin{equation}
R_{\mathrm{eff}}
=
\frac{\left(\sum_{t=1}^{N} w_t\right)^2}
{N \sum_{t=1}^{N} w_t^2}.
\end{equation}
where $N$ is the number of response tokens. A larger $R_{\mathrm{eff}}$ indicates that supervision is distributed more uniformly across tokens, with uniform SFT corresponding to $R_{\mathrm{eff}}=1$.

\begin{table}[htbp]
\centering
\small
\caption{Effect of margin and bandwidth on TrimSFT token weighting for Qwen2.5-Math-1.5B. Avg. denotes average@8 accuracy across the five mathematical reasoning benchmarks.}
\label{tab:weight-diagnostic}
\resizebox{\columnwidth}{!}{
\begin{tabular}{lccc}
\toprule
\textbf{$(m,\tau)$} &
\textbf{High-Weight (\%)} &
\textbf{$R_{\mathrm{eff}}$} &
\textbf{Avg.} \\
\midrule
$(1,1)$ & 14.92 & 0.212 & 30.58 \\
$(3,1)$ & 14.79 & 0.223 & 30.03 \\
$(6,1)$ & 15.56 & 0.236 & 31.61 \\
\addlinespace[2pt]
$(1,3)$ & 38.47 & 0.517 & 23.36 \\
$(1,6)$ & 66.42 & 0.831 & 18.11 \\
\addlinespace[2pt]
$(6,3)$ & 47.34 & 0.666 & 29.79 \\
$(6,6)$ & 86.34 & 0.923 & 24.18 \\
\midrule
SFT & 100.00 & 1.000 & 15.47 \\
\bottomrule
\end{tabular}
}
\end{table}

Table~\ref{tab:weight-diagnostic} shows that varying $m$ from 1 to 6 with $\tau=1$ leaves the fraction of high-weight tokens nearly unchanged, while increasing $\tau$ substantially broadens the weighted region and moves $R_{\mathrm{eff}}$ toward uniform SFT. These results provide additional evidence that $m$ mainly shifts the location of the weighting band, whereas $\tau$ controls its selectivity and trimming strength.

\section{Half-Trim Variants}
\label{app:variants}

Table~\ref{tab:variants-complete} reports additional results for TrimSFT and its two half-trim variants, TrimSFT-E and TrimSFT-H, on Qwen2.5-Math-1.5B. The main text reports representative margins \(m\in\{1,3,5\}\), while this appendix includes the full set of margins \(m\in\{1,2,3,4,5\}\). All variants are evaluated with fixed
\(\tau=1\).

\begin{table}[htbp]
\centering
\small
\caption{Comparison between TrimSFT and its half-trim variants with fixed \(\tau=1\). \textbf{Bold} marks the best result among the three variants.}
\label{tab:variants-complete}
\resizebox{\columnwidth}{!}{
\begin{tabular}{llccc}
\toprule
\textbf{$m$} & \textbf{Method} & \textbf{MATH500} & \textbf{AMC} & \textbf{AIME24} \\
\midrule
\multirow{3}{*}{1}
& TrimSFT   & 62.80 & 31.74 & \textbf{12.07} \\
& TrimSFT-E & \textbf{65.98} & \textbf{33.01} & 7.51 \\
& TrimSFT-H & 28.75 & 8.84 & 0.41 \\
\midrule

\multirow{3}{*}{2}
& TrimSFT   & 62.25 & 27.36 & 7.09 \\
& TrimSFT-E & \textbf{65.32} & \textbf{32.06} & \textbf{7.10} \\
& TrimSFT-H & 37.05 & 11.18 & 0.41 \\
\midrule

\multirow{3}{*}{3}
& TrimSFT   & \textbf{65.98} & \textbf{30.29} & \textbf{7.50} \\
& TrimSFT-E & 64.30 & 26.18 & 4.56 \\
& TrimSFT-H & 40.92 & 13.39 & 2.06 \\
\midrule

\multirow{3}{*}{4}
& TrimSFT   & 63.60 & \textbf{27.51} & \textbf{8.83} \\
& TrimSFT-E & \textbf{64.58} & 27.49 & 7.51 \\
& TrimSFT-H & 42.43 & 13.98 & 2.47 \\
\midrule

\multirow{3}{*}{5}
& TrimSFT   & 65.96 & 31.03 & \textbf{7.50} \\
& TrimSFT-E & \textbf{66.67} & \textbf{32.19} & 6.25 \\
& TrimSFT-H & 42.50 & 16.03 & 1.65 \\
\bottomrule
\end{tabular}
}
\end{table}

The results are consistent with the trends discussed in the main text.
TrimSFT-E is often competitive with TrimSFT on relatively easier or
medium-difficulty benchmarks such as MATH500 and AMC, suggesting that
preserving full supervision on high-gap tokens can sometimes be useful.
However, TrimSFT achieves better results on the harder AIME24 benchmark
across the reported margins. In contrast, TrimSFT-H consistently
performs much worse than both TrimSFT and TrimSFT-E, indicating that
preserving full supervision on low-gap tokens while trimming high-gap
tokens leads to an inferior trade-off. Overall, these results support
the benefit of two-sided trimming over relying on either half-trim
variant alone.

\section{Gradient-Mass Analysis}
\label{app:gradient-mass}

To further examine the asymmetric behavior of the half-trim variants, we analyze how each weighting scheme redistributes the token-level optimization signal. We conduct this analysis on 100 randomly sampled training examples from NuminaMath-CoT, covering 40,082 response tokens, using Qwen2.5-Math-1.5B. Following the setting in Section~\ref{Asym-variants}, we fix $\tau=1$ and consider $m\in\{1,3,5\}$. For each response token, we compute the weighted logit-gradient magnitude
\begin{equation}
g_t
=
w_t
\left\|
\operatorname{softmax}(z_t)-\operatorname{onehot}(y_t)
\right\|_2,
\end{equation}
and report the fraction of total gradient mass assigned to different logit-gap regions. For each $(m,\tau)$ setting, we define the low-, middle-, and high-gap regions as
$\Delta_t < m-\tau$,
$m-\tau \leq \Delta_t \leq m+\tau$,
and $\Delta_t > m+\tau$, respectively.

\begin{table}[htbp]
\centering
\small
\caption{Distribution of weighted gradient mass (\%) across logit-gap regions for TrimSFT and its half-trim variants on Qwen2.5-Math-1.5B. All settings use $\tau=1$.}
\label{tab:gradient-mass}
\resizebox{\columnwidth}{!}{
\begin{tabular}{llccc}
\toprule
\textbf{$m$} &
\textbf{Method} &
\textbf{Low} &
\textbf{Middle} &
\textbf{High} \\
\midrule
\multirow{3}{*}{1}
& TrimSFT   & 22.18 & 73.14 & 4.68 \\
& TrimSFT-E & 18.41 & 64.05 & 17.54 \\
& TrimSFT-H & \textbf{71.78} & 26.71 & 1.50 \\
\midrule
\multirow{3}{*}{3}
& TrimSFT   & 39.36 & 58.07 & 2.57 \\
& TrimSFT-E & 35.36 & 54.36 & 10.28 \\
& TrimSFT-H & \textbf{93.80} & 5.97 & 0.23 \\
\midrule
\multirow{3}{*}{5}
& TrimSFT   & 45.00 & 52.60 & 2.39 \\
& TrimSFT-E & 40.91 & 49.75 & 9.34 \\
& TrimSFT-H & \textbf{99.04} & 0.93 & 0.04 \\
\bottomrule
\end{tabular}
}
\end{table}

Table~\ref{tab:gradient-mass} reveals a clear asymmetry between the two sides of the weighting profile. TrimSFT-H, which preserves full supervision on low-gap tokens, concentrates most of its gradient mass in the low-gap region, reaching 71.78\%, 93.80\%, and 99.04\% as $m$ increases. In contrast, both TrimSFT and TrimSFT-E substantially suppress this region and maintain a larger share of the optimization signal in the intermediate region. Full TrimSFT further reduces the contribution of high-gap tokens relative to TrimSFT-E, providing an additional regularization effect on already well-separated tokens. These results support the interpretation that low-gap trimming is the primary source of stability, while high-gap trimming provides a complementary regularization effect.

\section{Token-Category Analysis}
\label{app:token-category}

Following the same experimental setting as Section~\ref{app:gradient-mass}, we conduct the token-category analysis on 100 randomly sampled training examples from NuminaMath-CoT using Qwen2.5-Math-1.5B, covering 40,082 response tokens. We group the decoded response tokens into four lightweight categories and compare standard SFT with TrimSFT under $(m,\tau)=(1,1)$.

\begin{table}[htbp]
\centering
\small
\setlength{\tabcolsep}{3.5pt}
\caption{Token-category analysis on Qwen2.5-Math-1.5B. Num., MS, and RC denote numeric tokens, mathematical symbols, and reasoning connectors, respectively. Ratio is the token fraction, Weight is the average TrimSFT weight, Mid. is the intermediate-gap fraction, and Grad. is the gradient-mass fraction.}
\label{tab:token-category}
\begin{tabular}{lccccc}
\toprule
\textbf{Type} &
\textbf{Ratio} &
\textbf{Weight} &
\textbf{Mid.} &
\textbf{SFT Grad.} &
\textbf{Trim Grad.} \\
\midrule
Num.  & 15.39 & 0.132 & 11.25 & 5.55  & 10.31 \\
MS    & 22.49 & 0.191 & 16.31 & 15.76 & 22.78 \\
RC    & 0.82  & 0.053 & 3.98  & 2.55  & 0.65 \\
Other & 61.30 & 0.158 & 13.06 & 76.14 & 66.25 \\
\bottomrule
\end{tabular}
\end{table}

Table~\ref{tab:token-category} shows that TrimSFT assigns a larger share of gradient mass to numeric and mathematical-symbol tokens than standard SFT, while reducing the contribution of the broad \textit{Other} category. In particular, the gradient-mass share increases from 5.55\% to 10.31\% for numeric tokens and from 15.76\% to 22.78\% for mathematical symbols. These results suggest that the reweighted optimization signal remains closely associated with math-relevant token categories rather than being concentrated primarily on generic response tokens.

\end{document}